\PassOptionsToPackage{numbers, compress}{natbib}
\documentclass{article}

\usepackage[preprint]{neurips_2019}

\usepackage{graphicx}
\usepackage{caption}
\usepackage{subcaption}
\usepackage{adjustbox}
\usepackage{float}
\usepackage[utf8]{inputenc} % allow utf-8 input
\usepackage[T1]{fontenc}    % use 8-bit T1 fonts
\usepackage{hyperref}       % hyperlinks
\usepackage{url}            % simple URL typesetting
\usepackage{booktabs}       % professional-quality tables
\usepackage{amsfonts}       % blackboard math symbols
\usepackage{nicefrac}       % compact symbols for 1/2, etc.
\usepackage{microtype}      % microtypography
\usepackage{amsmath}        % for multiline division
\usepackage[colorinlistoftodos]{todonotes}
\hypersetup{colorlinks,citecolor=black,linkcolor=black,urlcolor=black}
\usepackage{colortbl}
\usepackage{soul} 

\title{Prior Activation Distribution (PAD): A Versatile Representation to Utilize DNN Hidden Units}

\author{
  Lakmal Meegahapola\\
  EPFL \& Idiap Research Institute\\
  Switzerland\\
  \texttt{lakmal.meegahapola@epfl.ch} \\
  \And
  Vengateswaran Subramaniam\\
  Singapore Management University\\
  Singapore\\
  \texttt{vengates@smu.edu.sg} \\
  \And
  Lance Kaplan\\
  U.S. Army Research Laboratory\\
  USA\\
  \texttt{lance.m.kaplan.civ@mail.mil} \\
  \And
  Archan Misra\\
  Singapore Management University\\
  Singapore\\
  \texttt{archanm@smu.edu.sg} \\
}

\begin{document}

\maketitle

\begin{abstract}
 In this paper, we introduce the concept of Prior Activation Distribution (PAD) as a versatile and general technique to capture the typical activation patterns of hidden layer units of a Deep Neural Network used for classification tasks. We show that the combined neural activations of such a hidden layer have class-specific distributional properties, and then define multiple statistical measures to compute how far a test sample's activations deviate from such distributions. Using a variety of benchmark datasets (including MNIST, CIFAR10, Fashion-MNIST \& notMNIST), we show how such PAD-based measures can be used, independent of any training technique, to (a) derive fine-grained uncertainty estimates for inferences; (b) provide inferencing accuracy competitive with alternatives that require execution of the full pipeline, and (c) reliably isolate out-of-distribution test samples. 
  % and to harness value-added information from a Deep Neural Network. We demonstrate that (1) PADs leverage the behavior of hidden unit activations and this property can be utilized to provide accurate classification inferences (2) PADs provide predictive uncertainty estimates for inferences (3) PADs are capable of distinguishing out-of-distribution samples. We demonstrate that this simple, but unique technique can be used with a variety of deep learning models, hidden-layers types with no modification during the training phase. We used MNIST, CIFAR10, Fashion-MNIST, notMNIST datasets to characterize and evaluate the proposed technique. As a whole, this paper introduces a novel intuition regarding the behavior of hidden-unit activations in deep neural networks and we hope it would allow to grow a new family of algorithms around currently used deep neural network architectures.

%\textcolor{red}{\textbf{Vengat : Linear Separability ?, Intuition to DNNs ?}}

%and can be adapted to currently available models with minimum effort which is promising in terms of incorporating uncertainty estimates to critical real world applications
\end{abstract}

\section{Introduction}\label{sec:introduction}
Deep Neural Networks (DNNs) \cite{LeCunn2015} have rapidly become an indispensable mechanism for implementing machine intelligence for a variety of tasks, such as medical image analysis~\cite{Litjens2017}, chatbots for conversational interactions~\cite{Io2017} and navigation of autonomous vehicles \& robots~\cite{Pal2019,Pinto2016,Duan2016}.  DNNs represent state-of-the-art techniques for multi-class classification problems, which conventionally use the point estimates of the final softmax layer to identify the class with the highest confidence value. 

DNNs are still largely viewed as ``black box" models that generate inferences---a significant amount of ongoing research focuses on improving their \emph{final-layer} accuracy, often by increasing the depth of the inferencing pipeline (e.g., Resnet~\cite{He2015} with 152 layers). With a few notable exceptions (e.g.,~\cite{Alain2017,Yosinski2014,Yeh2018}), researchers have typically not devoted much systematic attention to characterizing or exploiting the activation values of intermediate, \emph{hidden} layers. In this work, inspired by the work of Alain \& Bengio \cite{Alain2017} in understanding hidden DNN layers, we ``unpack'' this black box and propose the novel concept of \emph{Prior Activation Distribution} (\textbf{PAD}) as a fundamental construct for characterizing DNNs.  PAD specifically focuses on the \emph{activation values} associated with hidden units (e.g., dense neurons, flattened convolutional layers, pooling layer values), and uses \emph{aggregated, statistical} properties of such activation values as a formal mechanism to tackle a variety of DNN-related problems.

We initially developed the PAD construct to quantify the \emph{predictive uncertainty} associated with DNN inferences. It is known that the confidence values of the softmax layer alone do not capture the uncertainty of the underlying inferencing process~\cite{Nguyen2015, Gal2016,AiOTA2019}. Recently proposed Bayesian Deep Learning (BDL) approaches \cite{MacKay1995,MacKay1992} can model such DNN uncertainty in a more theoretically-grounded manner, but impose significant computational complexity in both training and inference \cite{Gal2016}. Moreover, softmax-based inferences require the execution of the entire DNN pipeline, which may impose high latency when executed on resource-limited embedded platforms. We shall show that PAD serves as a \emph{versatile, computationally-inexpensive} way to quantify such uncertainty: PAD makes no assumption on the training mechanism and can be applied independent of the choice of regularization techniques (e.g., dropout,batch normalization, data augmentation). In addition, we provide evidence that (a) PAD-based uncertainty measures may enable more reliable filtering of out-of-distribution (OOD) data without compromising the base classification accuracy, and (b) the use of PADs may enable us to achieve competitive accuracy while only partially executing the DNN pipeline.

\subsection{Hypothesis and Contribution}\label{subsec:intuition}
Our hypothesis is that, given any existing (trained) DNN model, the activation values of each hidden unit of a DNN contain \emph{latent} information, that makes it more or less likely to be generated by a member of a specific class. By collecting the activation values from all training instances of a specific class, we can then create an appropriate, \emph{per-class}, representation of the \emph{typical range, or distribution} of each neuron's values. When making inferences (on a test sample), we posit that the larger the deviation of a hidden unit's activation value from this typical range, for a specific class, the lower the likelihood that the sample belongs to this class. By aggregating such deviation scores (through appropriate statistical features) across all the neurons in an hidden layer, we believe that we can better quantify the test sample's likelihood of belonging to this class. Overall, PADs allow us to analyze DNNs by understanding the behavior of  neurons in hidden layers,  which we believe represents a step towards the goals of making deep learning models more uncertainty-aware, less computationally complex and more interpretable. PADs also provide attestation to our belief that exploiting the behavior of hidden layers can help build richer models of DNN behavior than possible solely from output layer observations.

\noindent \textbf{Key Contributions:} We highlight the following key contributions:
\begin{itemize}
    \item We introduce the novel concept of Prior Activation Distribution (PAD), a simple technique to model hidden-unit activations of a DNN in multi-class classification problems. Further, we empirically demonstrate that PADs can be utilized to model different types of layers, regardless of the model architecture or regularization  techniques used. We also develop statistical measures, over PAD values, that help represent such hidden unit behavior. 
    \item We empirically demonstrate that PADs can capture and quantify the "Predictive Uncertainty" associated with a classification output. PAD-based uncertainty measures corrrespond closely to alternative, more complex models for uncertainty computation. 
    \item We show that, by using additional PAD-based features in conjunction with conventional output confidence scores, DNN classifiers can robustly identify and discard out-of-distribution (OOD) test samples, without sacrificing the ability to reliably classify in-distribution samples. We also provide early empirical evidence that PADs can be leveraged on to provide high classification accuracy, without executing the entirety of a DNN pipeline.
\end{itemize}

\section{Proposed Approach}\label{sec:approach}
\subsection{Formulating the Hidden Units of Hidden Layers}
Let's consider a trained DNN model $\textbf{G}$ for a $\textbf{|C|}$-class classification problem, which has been trained using training data $\textbf{X}_{train} = \{\textbf{x}_1,\textbf{x}_2, .. , \textbf{x}_m\}$ and training labels $\textbf{Y}_{train} = \{\textbf{y}_1,\textbf{y}_2, .. ,\textbf{y}_m\}$ where $m$ is the training dataset size and $C$ denotes the set of class labels.  Let $\textbf{Y}_{train}^{out} = \{\textbf{y}_1^{out}, \textbf{y}_2^{out}, .. ,\textbf{y}_m^{out}\}$ be the set of output labels, when evaluated on the $\textbf{X}_{train}$ it self. The convention is to calculate training accuracy using the instances where $\textbf{y}_i^{out}=\textbf{y}_i$ where $\textbf{y}_i^{out} \in \textbf{Y}_{train}^{out}$,  $\textbf{y}_i \in \textbf{Y}_{train}$ and $i \in [1,\textbf{m}]$.

Consider a $\textbf{G}$, with a set of layers $\textbf{L} = \{\textbf{l}_1,\textbf{l}_2, .. ,\textbf{l}_p\}$, such that the number of hidden units in each layer be denoted by the set $\textbf{S} = \{\textbf{s}_1,\textbf{s}_2, .. , \textbf{s}_p\}$, where $p$ = number of hidden layers. We can represent activation of a hidden unit $\textbf{a}_i$ in a particular layer $\textbf{l}_j$ as $\alpha_{\textbf{a}_i}^{\textbf{l}_j}$ where  $\textbf{l}_j \in \textbf{L}$, $\textbf{a}_i\in [1,\textbf{s}_j]$, $\textbf{s}_j \in \textbf{S}$. Here, $\textbf{s}_{j}$ represents the number of hidden units in the layer $\textbf{l}_j$. To avoid ambiguity, we positionally index layers from beginning to the end (thus $\textbf{l}_1$ represents the input layer),  and the hidden-units from top to bottom (thus $a_1^{\textbf{l}_j}$ represents the top-most neuron in the $j^{th}$ layer).

Extending this terminology, we define $\alpha_{\textbf{a}_j,\textbf{c}_t}^{\textbf{l}_i,\textbf{x}_e}$ where activation of a hidden unit with  (1) positional index $\textbf{a}_j \in [1,\textbf{s}_j]$ in layer $\textbf{l}_i \in \textbf{L}$, when (2) the input to $\textbf{G}$ is $\textbf{x}_e \in \textbf{X}_{train}$, and (3) output of the network is \textbf{correct}, and (4) it belongs to class $\textbf{c}_t \in \textbf{C}$. Here $\textbf{C} = \{\textbf{c}_1, \textbf{c}_2, .. , \textbf{c}_n\}$ is the set of classification outputs. When $\textbf{X}_{train}$ is used, with 1 stochastic forward pass of each $\textbf{x}_e \in \textbf{X}_{train}$ through model $\textbf{G}$, using the above definitions, we are able to obtain a distribution of activations, \emph{for each class, for each hidden unit}.  We refer to this set of distributions, as the \textit{Prior Activation Distribution} (\textbf{PAD}). 

We make the following two important observations: (a) \emph{Independent of Learning Technique:} The PAD distributions are derived merely by passing the elements of the training dataset through an \emph{already trained} DNN---the definition of PAD is thus agnostic to the choice of training methods and parameters; (b) \emph{Utilizes Correct Classifications only:} Only training instances that are correctly classified contribute to the PAD model. This makes intuitive sense: PAD is used to represent the distribution of neural behavior observed, per class, only when the model is accurate.

For a hidden unit in the positional index of $\textbf{a}$ in layer $\textbf{l}$, \textbf{PAD} can be denoted using the notation;
\begin{equation}\label{eqn:pad}
    \textbf{PAD} = \{ \textbf{D}(\mu_{a,c_1}^l,\sigma_{a,c_1}^l), \textbf{D}(\mu_{a,c_2}^l,\sigma_{a,c_2}^l), ... , \textbf{D}(\mu_{a,c_n}^l,\sigma_{a,c_n}^l) \}
\end{equation}

\begin{equation}\label{eqn:mean}
     \mu_{a,c_t}^l = \frac{\sum\alpha_{a,c_t}^{l,x_e}}{\text{count}_{c_t}}\; \; \mbox{and}\;
     \sigma_{a,c_t}^l = \sqrt{\frac{\sum(\alpha_{a,c_t}^{l,x_e} - \mu_{a,c_t}^l)^2}{\text{count}_{c_t}}}
     \text{ where } \textbf{c}_t \in \textbf{C} \text{, } \textbf{x}_e \in \textbf{X}_{train}
\end{equation}

%\begin{equation}\label{eqn:std}
 %   \sigma_{a,c_t}^l = \sqrt{\frac{\sum(\alpha_{a,c_t}^{l,x_e} - \mu_{a,c_t}^l)^2}{\text{count}_{c_t}}} %\text{ where } \textbf{c}_t \in \textbf{C} \text{, } \textbf{x}_e \in \textbf{X}_{train}
%\end{equation}

In this definition, $\textbf{D}$ denotes any arbitrary empirical distribution, $count_{\textbf{c}_t}$ is the number of accurate inferences for $\textbf{x}_e$ which outputs a particular class $\textbf{c}_t$. As suggested by our hypothesis, the above definitions allow us to model each hidden unit as a PAD which consists of several distributions, each of which characterize how the hidden-unit activations should behave to produce a particular classification ($\textbf{c}_t \in \textbf{C}$). Note also that we make no distributional assumptions (e.g., Gaussian, often used in prior work~\cite{Gal2016,Gal2016v2,Herzog2013,Nuzzo2014}) on $\textbf{D}$; in Section~\ref{subsec:activationbehavior}, we shall see that these values are, in fact, quite arbitrary.

\subsection{Inference Using PADs: KL-divergence Z-Score metrics}
We now describe how statistical properties of such distributions are used to evaluate the `fit to a specific class' of a test sample during the inferencing phase. After choosing a particular layer $\textbf{l}$ which we want to model with \textbf{PAD}s, we obtain PADs for all hidden-units in $\textbf{l}$ using the training dataset, denoted by $\textbf{priors}_l = \{\textbf{PAD}_1, \textbf{PAD}_2, ... , \textbf{PAD}_s\}$. During the inferencing process, the test sample $\textbf{input}_{test}$ is passed through the DNN and generates a set of $s_l$ activation values (one for each hidden unit), in addition to the output prediction (at the final softmax layer) by the DNN. Let us denote these activation values layer $\textbf{l}$ with $\textbf{s}$ hidden units as $\textbf{activations}_{test} = \{\gamma_1, \gamma_2, .. , \gamma_s\}$. We then propose the following 2 representative statistical features to capture the similarity (or divergence) between the activation firings represented by PAD and those resulting from the test instance: (a) the \emph{KL-score} feature looks at the activation values across all hidden units of a layer jointly, while (b) the \emph{Z-score} feature first measures per-hidden unit divergence in activation values before aggregating across all hidden units.

\subsubsection{KL-Score Metric}
At a high-level, the KL-Score considers the set of individual activation values of a layer as a whole--i.e., as a $s$ dimensional vector, and compares the test-instance vector against each of the $|C|$ PAD-based vectors. More specifically, for a layer $\textbf{l}$ with $\textbf{s}$ nodes, the PAD vector  for a class $c_t$ consists of $s$ elements, where the $a^{th}$ element is obtained by taking the \emph{mean} value of the activation values $\mu_{a,c_t}^l$. The distance between the test instance and class $c_t$ is computed by the KL-divergence of the normalized values of $\textbf{activations}_{test}$ and the activation vector for class $c_t$. In this fashion, one can compute the overall KL-divergence vector $klscores^l$, whose elements consist of the KL-divergence measure for each of the $n = |C|$ classes--i.e., 
$\label{eqn:klscores}
    \textbf{klscores}^{l} = 
    \{
    klscore_{c_1} ,
    klscore_{c_2} , 
    .... , 
    klscore_{c_n}
    \}.
$
%\end{equation}

Given this formulation, the higher the KL-score, the lower the likelihood of a test instance belong to that class. Accordingly, to classify the test sample using just the KL-score values at hidden layer $l$, we would generate an output corresponding to  $\min(\textbf{klscores}^l)$.

\subsubsection{Z-score Metric}
This approach first looks at each (hidden,class) individually and computes a Z-score\footnote{Technically, this is a \emph{pseudo} Z-score, at it does measure the distance of a data point (in terms of the number of standard deviations) from the mean, but does not assume a Normal distribution.}, representing the degree to which the test sample's activation value can be considered an outlier, given the representative mean ($\mu_{a,c_t}^l$) and standard deviation ($\sigma_{a,c_t}^l$). When $n = |C|$, this pseudoZ-score, across all classes, but for neuron $a$ in hidden layer $l$, is first computed as:
$
    \textbf{zscores}_{\textbf{a}}^{\textbf{l}} = 
    \{
    \frac{\gamma_a - \mu_{a,c_1}^l}{\sigma_{a,c_1}^l} ,
    \frac{\gamma_a - \mu_{a,c_2}^l}{\sigma_{a,c_2}^l} , 
    .... , 
    \frac{\gamma_a - \mu_{a,c_n}^l}{\sigma_{a,c_n}^l}
    \}.
$

Subsequently, the Z-score \textbf{$zscores^l$}, across all the $s_l$ neurons in layer $l$, is computed as the mean of these $s_l$ distinct values, defined as:
\begin{equation}\label{eqn:zscore_l}
\begin{split}
    \textbf{zscores}^{\textbf{l}} & = 
    \{
    \frac{\sum_{a=1}^s \frac{\gamma_a - \mu_{a,c_1}^l}{\sigma_{a,c_1}^l}}{s} ,
    \frac{\sum_{a=1}^s \frac{\gamma_a - \mu_{a,c_1}^l}{\sigma_{a,c_2}^l}}{s} , 
    .... , 
    \frac{\sum_{a=1}^s \frac{\gamma_a - \mu_{a,c_1}^l}{\sigma_{a,c_n}^l}}{s}
    \} \\ 
    & = 
    \{
    zscore_{c_1},
    zscore_{c_2},
    .... , 
    zscore_{c_n}
    \}
\end{split}
\end{equation}

Given this formulation, the \emph{higher} the Z-score, the lower the likelihood of a test instance belonging to that class. Accordingly, to classify the test sample using the observed activations at  hidden layer $l$, we would generate an output corresponding to  $\min(\textbf{zscores}^\textbf{l})$.

\section{Preliminary Analysis}\label{sec:preliminary}
%In section \ref{subsec:hiddenlayers}, we disclose some results from our preliminary analysis regarding hidden-unit activations which became the base for the intuition behind this work. Then, we describe how models are trained to obtain PADs in \ref{subsec:trainingphase}. Then we empirically demonstrate the methodology we use to obtain KL-scores (based on KL-divergence) and Z-scores and examine the characteristics of these values in \ref{subsec:klz_scores}. Using KL-scores and Z-scores of PADs, we demonstrate the power of using PADs by showing that (1) PADs provide classification inferences leveraging the linear separability of intermediate layers in \ref{subsec:inference}. (2) PADs give accurate and sensitive uncertainty estimates in \ref{subsec:uncertainty}. Next we present our findings regarding the behavior of PAD based KL-scores for out-of-distribution data in \ref{subsec:outofdist}. 

% In these experiments we demonstrate the PAD based modelling of intermediate layers can be done for (1) Different Architectures (2) Different types of layers (e.g. Conv2D, Dense, Pooling layers) in the above experiments (section \ref{subsec:inference}). 

 %During the Training Phase, each of model has been trained using the general training procedure with hyper parameter tuning.

\begin{table}[t]
\begin{small}
\begin{center}
\begin{adjustbox}{width=\textwidth,center}
        \begin{tabular}{c c c c c c} 
        \hline 
        Dataset   & Model Summary                                   &             Training Testing Split    & Regularization techniques          & Optimizer                                & Reference \\ \hline 
        \textbf{MNIST}     & C32, D128, D10      & 60000-10000                                           & Dropout                            & adam                                     & \textbf{MA1}       \\ 
        \textbf{MNIST}     & C64, C128, D128, D10 (Modified LeNet-5 \cite{LeCunn1998})& 60000-10000    &  Dropout                            & adam                                     & \textbf{MA2}       \\ 
        \textbf{MNIST}     & C32, C64, D128, D10 \cite{Keras2019_MNIST} & 60000-10000    & Dropout                            & adam                                     & \textbf{MA3}       \\ 
        \textbf{CIFAR10}   & C32, C32, C64, C64, D512, D10 \cite{Keras2019_CIFAR10}   & 50000-10000          & Dropout                            & adam                                     & \textbf{CA1}       \\ 
        \textbf{CIFAR10}   & Modified All Convolutional Net \cite{Springenberg2015,Kumar2018} & 50000-10000 & Batch Normalization, Dropout       & RMS(learning rate = 0.01, decay = 1e-6)  & \textbf{CA2}       \\ 
        \textbf{CIFAR10}   & All Convolutional Net \cite{Springenberg2015}    & 50000-10000               &  Data Augmentation                         & SGD(learning rate = 0.01, decay = 1e-6)  & \textbf{CA3}       \\
        \textbf{Fashion-MNIST}  & C64, C32, D256, D10 \cite{Reid2018} & 60000-10000 & Dropout & adam & \textbf{FMA1}  \\
        \textbf{NotMNIST}  & - & NA-100000 & - & - & \textbf{NM1}  \\ 
        \textbf{Modified-MNIST}  & - & NA-10000 & - & - & \textbf{MM1} \\ \hline
        \end{tabular}
        \end{adjustbox}
\end{center}
\end{small}
\vspace{0.05 in}
\caption{\small{Summary of the datasets and  model architectures. The notation uses C, D to represent Conv2D \& Dense units respectively, while the subsequent number specifies the number of hidden-units:e.g., D128 represents a layer with 128 dense neurons.}}
\label{tab:datasetsmodels}
\vspace{-0.2 in}
\end{table}

In this section and section~\ref{sec:experiments}, we extensively analyze the properties of PAD (and the related \emph{KL} and \emph{Z-score} features), using multiple benchmark classification datasets: MNIST \cite{LeCunn1998}, CIFAR10 \cite{Krizhevsky2009}, Fashion-MNIST \cite{xiao2017}, notMNIST \cite{Bulatov2011}, Modified-MNIST\footnote{Modified MNIST dataset was created by combining pairs of consecutive images of MNIST. A sample from this dataset is shown in \ref{fig:modifiedmnist}} datasets. All the experiments were implemented and evaluated using Python \cite{Python2019} with Keras library \cite{chollet2015} with a Tensorflow \cite{tensorflow2015} backend. All the model configurations we used are summarized in Table~\ref{tab:datasetsmodels}.
\begin{figure*}[t]
\begin{center}
    \begin{minipage}[t]{0.6\textwidth}
    \begin{subfigure}[b]{0.32\textwidth}
          \includegraphics[width=\textwidth, height=0.77 in]{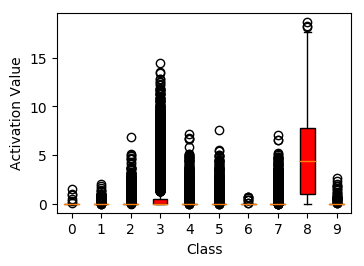}
          \caption{Hidden-unit \#1 \label{fig:1}}
     \end{subfigure}
    \hfill
    \begin{subfigure}[b]{0.32\textwidth}
      \includegraphics[width=\textwidth, height=0.77 in]{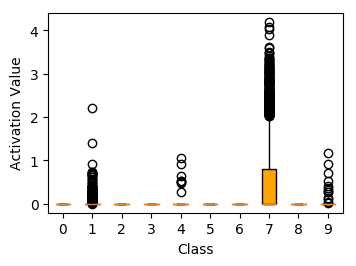}
      \caption{Hidden-unit \#65 \label{fig:65}}
    \end{subfigure}
    \hfill
    \begin{subfigure}[b]{0.32\textwidth}
      \includegraphics[width=\textwidth, height=0.77 in]{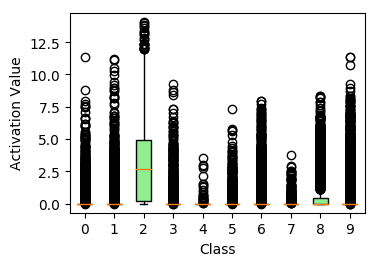}
      \caption{Hidden-unit \#128 \label{fig:128}}
    \end{subfigure}
    \hfill
    \caption{Examples for activation distributions of hidden-units of a DNN trained for MNIST dataset. The hidden-units here are from a layer with 128 neurons \& numbering was done from top to bottom.}
    \label{fig:activations}
    \end{minipage}
    \begin{minipage}[t]{0.39\textwidth}
    \vspace{-1.1 in}
    \centering
      \includegraphics[width=0.8\textwidth]{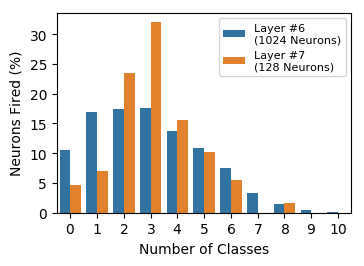}
      \caption{Histogram of number of classes against fired neurons percentage}
      \label{fig:histogram}
    \end{minipage}
\end{center}
\vspace{-0.3 in}
\end{figure*}

\subsection{Behavior of Hidden-Layer Activations of DNNs}\label{subsec:hiddenlayers}
\label{subsec:activationbehavior}
We carried out several preliminary experiments to understand the behavior of hidden layer activations. We will use the following example to illustrate our findings. We trained a DNN for MNIST dataset with configuration MA3 (\ref{tab:klz}) (a model with 1024 flattened values from Conv2D layers -- layer 6, 128 dense neurons -- layer 7 with ReLU activations). Figure~\ref{fig:activations} provides boxplot~\cite{Hunter2007} based visualizations (one for each of the 10 classes) of some representative dense neurons in layer 7, and Figure~\ref{fig:histogram} provides a histogram plot of the number of hidden units that fire for distinct classes (for both layer 6 \& 7). 

%\am{Lakmal--add a qualifier on what type of layers these are.}

We make the following observations. hidden unit number 1 (Figure~\ref{fig:1}) generates a wide range of activation values for the $9^{th}$ class, but has activation values very close to 0 for the $1^{st}$, $2^{nd}$ $7^{th}$ \& $10^{th}$ classes. On a similar note, hidden-unit 128 (the middle neuron in th layer, illustrated in Figure~\ref{fig:128}) shows a unique pattern for the activation value distribution for the $3^{rd}$ class. On the other hand,  hidden unit 65 (Figure~\ref{fig:65}) effectively does not generate non-zero activation values for 6 output classes. From similar analysis performed with both MNIST and CIFAR10 datasets, we observe that:  (a) hidden unit activations typically possess a unique distributional pattern for one or more classes and (b) the distributions are not necessarily normal. These unique patterns might help in both discriminating among classes and in quantifying uncertainty--e.g., if the activation value of hidden-unit 1 for an unknown test instance (that has been declared to be the $9^{th}$ class by the softmax output) is, say, 15.0 rather than 5.1 (closer to the mean of $9^{th}$ class), the DNN is likely to be more uncertain of this classification. Our plots clearly show that distributions are not typically normal. In addition, Figure~\ref{fig:histogram} plots the histogram (percentage) of hidden units, as a function of the number of distinct classes that activate each hidden unit at least once. We see that, across both layers 6 and 7, the dominant majority of hidden units are fired by three or fewer of the 10 classes. This result provides further evidence that most hidden units have distinct \emph{class-dependent} activation patterns, lending further credence to our exploration of PADs as a means for identifying class labels for test samples. 

%\am{Lakmal: I've put in text assuming that you will put in Vengat's graph for Figure 1(d). Do remember to also fix fig:XXX above.} 

%) which might allow us to differentiate between outputs of a DNN. For example, if the activation value of hidden-unit 1 for a particular input is 5.1 (closer to the mean of $9^{th}$ class in figure~\ref{fig:1}), it is highly likely that the output is closer to $9^{th}$ class than any other output class. %\textcolor{red}{\textbf{Vengat : Percentage of neurons exhibit this kind of behaviour (using kurtosis)?}} 
% \textcolor{red}{\textbf{Vengat : One observation from the graphs, in Hidden-unit #1, As the class 8 has more significance, class 3 has half of it. Similarly, in Hidden-unit #65, As class 7 is highly significant, next significant one is class 1.}} 

% \textcolor{red}{\textbf{Vengat : In the graph, Class - 0 to 9 ?}}

%\subsection{Training Phase}\label{subsec:trainingphase}
%During the Training Phase, each of the 6 models described in Table~\ref{tab:datasets} have been trained using the normal training procedure. We used standard procedures such as hyper-parameter tuning during the training process and various regularization techniques such as Dropout, Batch Normalization were used where and when necessary. After obtaining the optimal model for the architecture, we obtain PADs for each of these models using an additional epoch of passing training dataset through each trained model (see Table~\ref{tab:datasets} for model summaries).

\subsection{Characterizing KL-score and Z-Score}\label{subsec:klz_scores}
\begin{figure*}[t]
\begin{center}
    \begin{minipage}[t]{0.6\textwidth}
    \begin{subfigure}[b]{0.24\textwidth}
          \includegraphics[width=\textwidth]{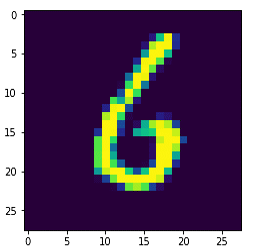}
          \caption{MNIST:\\"6" \label{fig:number6}}
     \end{subfigure}
    \hfill
    \begin{subfigure}[b]{0.24\textwidth}
      \includegraphics[width=\textwidth, height=0.77 in]{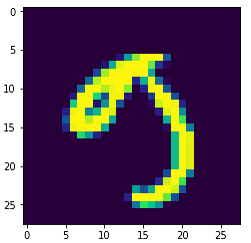}
      \caption{MNIST:\\"9" \label{fig:number5}}
    \end{subfigure}
    \hfill
    \begin{subfigure}[b]{0.25\textwidth}
      \includegraphics[width=\textwidth, height=0.77 in]{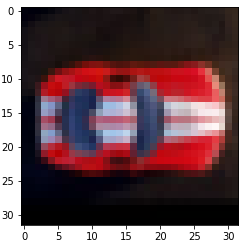}
      \caption{CIFAR10:\\"automobile"\label{fig:1_automobile}}
    \end{subfigure}
    \hfill
    \begin{subfigure}[b]{0.24\textwidth}
      \includegraphics[width=\textwidth, height=0.77 in]{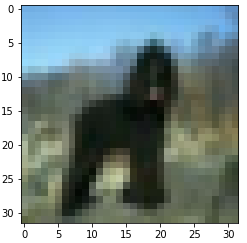}
      \caption{CIFAR10:\\"dog"\label{fig:5_dog}}
    \end{subfigure}
    \caption{Examples for PAD characterization}
    \label{fig:5graphs}
    \end{minipage}
    \begin{minipage}[t]{0.39\textwidth}
    \vspace{-1.13 in}
    \centering
      \includegraphics[width=0.45\textwidth]{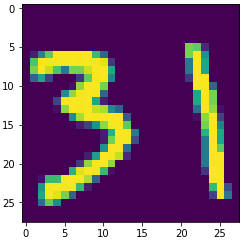}
      \caption{Example from Modified-MNIST dataset}
      \label{fig:modifiedmnist}
    \end{minipage}
\end{center}
\vspace{-0.3 in}
\end{figure*}

In this section, we evaluate the characteristics of KL-divergence and Z-score based values obtained for several images from MNIST and CIFAR10 datasets using MA3 \& CA1 configurations (Table~\ref{tab:datasetsmodels} \& \ref{tab:results}), respectively. Table~\ref{tab:klz} shows different KL and Z-score values, as well as the output of the softmax layer, the classes corresponding to the minimum KL and Z-scores and the ground truth, for 4 different representative images (2 for MNIST, 2 for CIFAR10) illustrated in Figure~\ref{fig:5graphs}.

In the MNIST sample (ground truth=6) in Figure~\ref{fig:number6}, the minimum KL score (1.073) and Z-score (0.594) for this sample (plotted in Table~\ref{tab:klz}) correspond to the correct class ``6". In this case, the softmax output, the minimum KL-score and the minimum Z-score label all agree and are correct. In contrast, for Figure~\ref{fig:number5}, the class with the minimum KL and Z-score is 9 (agreeing with the ground truth), whereas softmax output suggests the class 5. In this case, the PAD-related features provide a correct classification while the output softmax does not.
Figure~\ref{fig:1_automobile} depicts an interesting example where the two top softmax output candidates ("ship" with 76.5\% confidence and "truck" with 18.8\% confidence) are both incorrect. However, the KL and Z-score metrics provide ``automobile" (the correct inference) and ``truck" respectively. Further, in Figure~\ref{fig:5_dog}, the softmax layer outputs the class "truck" (confidence> 73.1\%), whereas the  KL and/or Z-scores correctly indicate that the output should be "dog". While we defer the presentation of comprehensive results on overall accuracy till Section~\ref{sec:experiments}, the examples presented here do attest to the discriminative potential of PADs.

%It should also be noted that for configurations MA1, CA1 \& CA3, EnsOR outperformed softmax output slightly. This demonstrate that incorporating and making use of hidden-layer information of DNNs allow fine tuning inferences for higher accuracies. 
%If we take KL-vaues into consideration and use our basic intuition of neuron activations and their behavior, the value 1.3362 which corresponds to the class "automobile" and the value "4.3732" which corresponds to the class "deer"suggest that given the architecture of the model, given the training data this model has seen, the hidden-unit neurons have behaved in a manner which is more similar to instances where neurons have behaved in the training when outputing class "automobile" as compared to class "deer". 

\begin{table}[t]
\begin{small}
\begin{center}
\begin{adjustbox}{width=\textwidth,center}
        \begin{tabular}{c c c c c c} 
        \hline
         Example               & KL-scores     & Softmax Output & min of KL & min of Z-scores & Ground truth \\
                        & Z-scores     &  &  &  &  \\
        \hline 
         Figure~\ref{fig:number6}
        & \{2.223, 1.627, 1.653, 1.599, 1.539, 1.364, \textbf{1.073}, 1.169, 1.181, 1.239\} 
        & 6 (98\%)
        & 6 
        & 6
        & 6\\ 
         & \{1.987, 1.147, 1.062, 0.975, 0.881, 0.733,\textbf{0.594}, 0.633, 0.636, 0.657\} & & & &  \\

         Figure~\ref{fig:number5}
        & \{2.445, 2.038, 2.08, 1.566, 1.442, 1.207, 1.269, 1.188, \textbf{1.183}, 1.125\}
        & 5 (78\%)
        & 9 
        & 9
        & 9\\ 
         & \{1.951, 1.321, 1.224, 0.877, 0.702, 0.58, 0.600, 0.567, \textbf{0.566}, 0.547\} & & & &  \\

         Figure~\ref{fig:1_automobile}
        & \{2.465, \textbf{1.335}, 3.798, 3.483, 4.373, 4.006, 3.849, 4.244, 1.396, 1.454\} 
        & ship (76.5\%)
        & automobile 
        & truck
        & automobile\\ 
        & \{1.951, 1.321, 1.224, 0.877, 0.702, 0.58, 0.600, 0.567,0.566,\textbf{0.547}\} \\

         Figure~\ref{fig:5_dog}
        & \{2.223, 1.627, 1.653, 1.599, 1.539, 1.364, \textbf{1.073}, 1.169, 1.181, 1.239\} 
        & truck (73.1\%)
        & dog 
        & dog
        & dog\\ 
         & \{1.987, 1.147, 1.062, 0.975, 0.881, 0.733, \textbf{0.594}, 0.633, 0.636, 0.657\} \\
        \hline 
        \end{tabular}
        \end{adjustbox}
\end{center}
\end{small}
\caption{Example figures and KL-score, Z-score characterization}
\label{tab:klz}
\vspace{-0.25 in}
\end{table}

\section{Experimental Evaluation of PAD Performance}\label{sec:experiments}
In this section, we empirically show that PADs can be used for a variety of uses, ranging from uncertainty quantification to ensuring highly accurate inferences.

% Results shown here for CA2 contains results from flattened values from Conv2D layer (24th layer) close to softmax output

\subsection{Using PADs for Uncertainty Quantification}\label{subsec:uncertainty}

\begin{figure*}[t]
\begin{center}
     \begin{minipage}[t]{0.38\textwidth}
     \centering
          \includegraphics[width=0.99\textwidth]{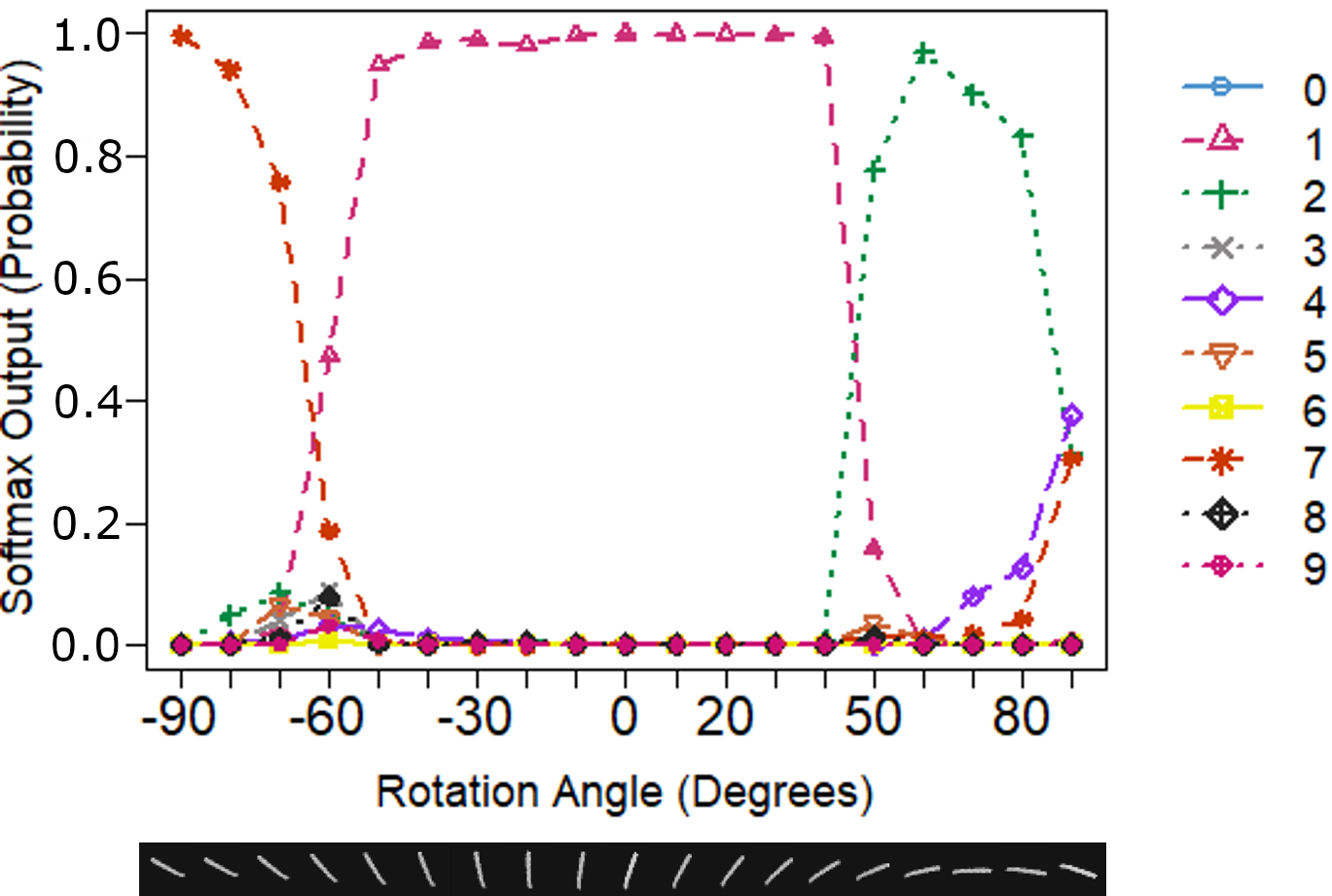}
          \caption{Softmax (MA1 model): Rotational MNIST}  \label{fig:softmax_outputs}
     \end{minipage}
    \hfill
    \begin{minipage}[t]{0.38\textwidth}
    \centering
      \includegraphics[width=0.99\textwidth]{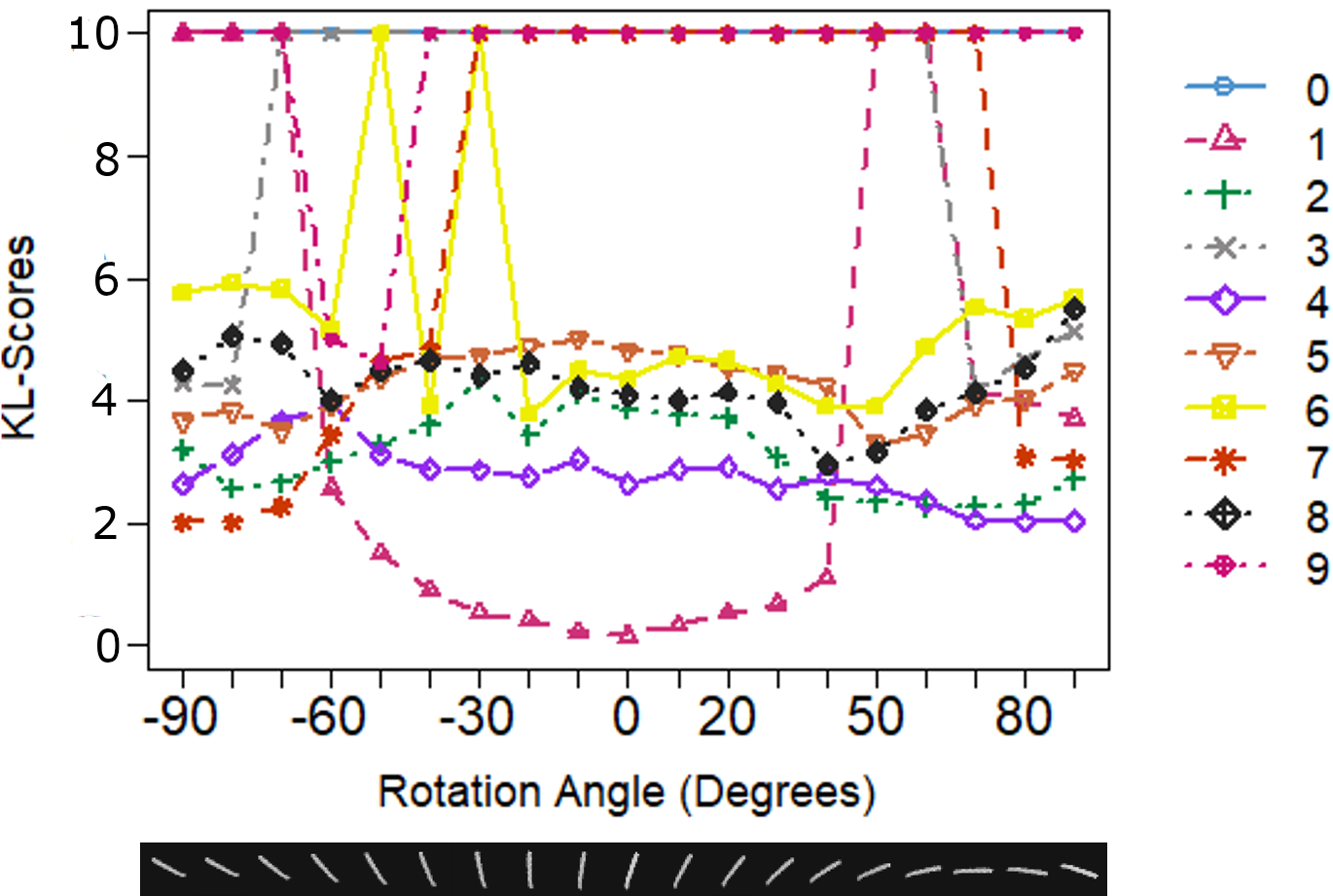}
      \caption{KL-Score (MA1 model):  Roational MNIST}
      \label{fig:kl_outputs}
    \end{minipage}
    \hfill
    \begin{minipage}[t]{0.22\textwidth}
    \vspace{-1.4 in}
    \centering
    \resizebox{\textwidth}{!}{
    \begin{small}
    \begin{tabular}{c c c} 
                    \hline
                     & MNIST & CIFAR-5 \\
                    \hline \hline
                    MC Dropout \cite{Gal2016v2}        & 99.5 \% & 84 \%  \\ \hline
                    Deep Ensemble \cite{Lakshminarayanan2017}  & 99.3 \% & 79 \%  \\ \hline
                    EDL \cite{Sensoy2018}          & 99.3 \% & 83 \%  \\ \hline
                    Softmax     & 99.4 \% & 80 \%   \\ \hline
                    KL      & 99.1 \% & 79 \%  \\ \hline
                    Z-score        & 96.3 \% & 76 \% \\
                    \hline 
        \end{tabular}
    \end{small}}
    %\vspace{0.29 in}
    \captionof{table}{Comparison of classification accuracies for MNIST and CIFAR-5 for uncertainty quantifying techniques using LeNet configuration in \cite{Louizos2017}}
    \label{tab:comparison}
    \end{minipage}

\end{center}
\vspace{-0.3in}
\end{figure*}

In this section, we observe how uncertainty is quantified using PADs of hidden layer units. Similar to experiments carried out by Gal \textit{et. al.} \cite{Gal2016v2} and Sensoy \textit{et. al.}\cite{Sensoy2018}, we carried out experiments using PADs (with several configurations) on MNIST images under varying degrees of rotation. Figures~\ref{fig:softmax_outputs} and ~\ref{fig:kl_outputs} show the softmax outputs and KL-scores obtained using the MA1 model (Table~\ref{tab:datasetsmodels}), using rotations on an original ``1" sample. For rotation angles between (-90$^\circ$,-65$^\circ$), both softmax output and KL-scores suggest that the output is in fact 7. But, the softmax output gives confidence values in excess of 75\% (more than 90\% for certain angles), which is over-optimistic given that the model was never trained for such images. Similar results were observed in Gal \textit{et. al.}, where they obtain a distribution of softmax outputs using dropout. However, even the dropout-based technique (as well as prior Bayesian approaches) result in higher confidence values. The PAD-based approach (Figure~\ref{fig:kl_outputs}), however provides a more conservative picture: while the class ``7" does have the lowest KL-score, the KL-scores of other classes (e.g., ``2" \& ``4") are quite similar, indicating that the DNN is not very confident of its inference. Similar results hold for rotation angles between (50$^\circ$,90$^\circ$), where the softmax output continues to have a misleadingly high confidence ($>$75\%). However, in the ``normal range" of (-40$^\circ$,40$^\circ$), the KL-score for the correct class (``1") is \emph{significantly lower} than that of other classes, indicating that the DNN has low inferencing uncertainty. In addition, we see that the PAD-based KL-score is able to offer a \emph{finer-grained} measure for uncertainty, with the KL-score for ``1" increasing gradually, even for 5$^\circ$ increments. In contrast, the softmax output remains high ($>$96\% for all rotations between (-60$^\circ$,50$^\circ$). Further, Table~\ref{tab:comparison} compares classification accuracies of models and techniques which quantify different types of uncertainty (it should be noted that the purpose of these techniques is not to have high accuracies per se, but to have reliable uncertainty estimates).  

%Another important characteristic of the PAD based approach is the sensitivity it has for these rotating images. As seen from angles from -60 to 0 degrees, while the image is been rotated to the usual position, the KL-score for output 1, decreases gradually. It's clearly seen the KL-scores are sensitive to even 5 degree rotations unlike softmax outputs which provide over 96\% confidence for all rotation angles between -60 and 50 degrees. Our methodology clearly differentiates from Sensoy \textit{et. al.}'s work \cite{Sensoy2018} because, their approach just outputs that the DNN is uncertain for images in a particular range of rotations with near 100\% confidence where as PAD based approach provides the nearest possible answer, with an uncertainty estimate. 

\subsection{Using PADs for Inference} \label{subsec:inference}
We now show how the discriminative capabilities of KL-score and Z-score values (illustrated in  Section~\ref{subsec:klz_scores}) can help improve the inferencing process. To compare with the baseline approach (based on the softmax output layer), we consider several alternative PAD-based inferencing strategies which operate on the hidden-layer activation values: \textit{KL} and \textit{Z-score} approaches output the class with the lowest KL-Score and Z-score, respectively; \textit{EnsAND} generates a class label only when all 3 measures (softmax, KL, Z-score) unanimously agree on the same class; while \textit{EnsOPT} serves as an alternative optimal (\emph{oracular}) baseline that picks the correct class if at least one of the 3 approaches (softmax, KL, Z-score) is correct. 

Table~\ref{tab:results} plots the classification accuracy of these approaches (for different datasets and models). We see that PAD-based \textit{KL} and \textit{Z-score} accuracy comparable to the softmax baseline, especially when applied to neural activations in the latter (deeper) part of the DNN. Further, an optimal ensemble technique \emph{EnsOPT}, which smartly combines the PAD and softmax output inferences, can in fact exceed baseline accuracy, at least for the MNIST, CIFAR10 and Fashion-MNIST benchmark datasets.  
We additionally considered configurations CA2, CA3 and observed that PAD-based accuracy increases as we go deeper in the network \footnote{CA2 configuration used flattened values ($24^{th}$ layer), activation matrices of conv2d layers ($12^{th}$,$17^{th}$ and $20^{th}$ layers) to formulate PADs while CA3's pooling layers were used. In CA2, Conv2D layers are of the shapes (8,8,128), (16,16,64) and had 8192, 16384 individual values which we considered as hidden units to create PADs for layers in the middle of the network.}--for example, in CA2, Z-score based accuracy was 53.1\%, 69.4\%, 83.45\%, 84.05\% for convolutional layers numbered 12,17,20 and 24 respectively. This result is consistent with prior work (Section~\ref{sec:relatedwork}) which demonstrates that deeper layers of a DNN are able to capture more specific features. While we omit results due to space limitations, we also tested PAD-based inferencing using other types of hidden layer--e.g., flattened values in CA2, pooling layers in CA3, etc., as well as when different regularization techniques (e.g., Data Augmentation, Dropout, Batch Normalization) were used. The results are consistent: PAD-based classification provides high accuracy under partial computation in all cases, demonstrating the  \emph{versatility} of this representation. 

As a final illustration of using PAD in inferencing, we consider the use of of DNNs in mission-critical scenarios, where we desire that automated DNN classification should have `near-100\%" accuracy--i.e., it should aggressively refer \emph{uncertain} test samples to explicit manual verification. (An example would be DNNs used in medical image analysis)  There is clearly a tradeoff between \emph{coverage} (the percentage of samples that the DNN automatically classifies) and \emph{accuracy} (defined over the \emph{covered} samples). We compare three alternatives in terms of this tradeoff: (a) baseline, which uses the softmax \emph{confidence} value to quantify uncertainty and thus invokes manual intervention when this confidence falls below a threshold; (b)  KL-score and (c) Z-score, both of which invoke manual intervention when the corresponding metric exceeds a specified threshold. Figure~\ref{fig:mnist_fmnist} plots the resulting accuracy vs. coverage tradeoff. We see that the KL-score approach is able to explore this tradeoff continuum--e.g., it can ensure over 99.99\% accuracy by filtering out around 20\% of the test samples for manual inspection. In contrast, while the baseline softmax approach does have high initial accuracy, it cannot easily push the accuracy higher as confidence does not reliably indicate uncertainty.

\begin{table}[t]
\begin{small}
\begin{center}
\begin{adjustbox}{width=\columnwidth,center}
        \begin{tabular}{c c c c c c c} 
        \hline
        Reference (Table~\ref{tab:datasetsmodels}) & Layer     & \multicolumn{5}{c}{Accuracy} \\
                                             &  & Softmax & KL      & Zscore  & EnsAND  & EnsOPT   \\ \hline 
        \textbf{MA1}  & 128-Dense Layer                                   & 99.03\% & 98.40\% & 98.27\%  & 97.88\% & 99.34\%  \\ 
        \textbf{MA2}  & 128-Dense Layer                                     & 99.44\% & 99.17\% & 99.04\%  & 98.82\% & 99.59\%  \\ 
        \textbf{MA3}  & 128-Dense Layer                                      & 99.56\% & 99.31\% & 99.15\%  & 99.08\% & 99.66\%  \\ 
        \textbf{CA1}  & 512-Dense Layer                                   & 88.17\% & 88.11\% & 88.24\%  & 86.85\% & 89.63\%  \\ 
        \textbf{CA2}  & 2048-Flattened Value Layer ($24^{th}$ Layer)  & 89.31\% & 84.11\% & 84.05\%  & 79.25\% & 92.95\%  \\ 
        \textbf{CA3}  & Average Pooling Layer                       & 92.75\% & 89.11\% & 85.74\%  & 78.31\% & 94.97\%  \\ 
        \textbf{FMA1}  & 256-Dense Layer                       & 91.81\% & 91.22\% &91.32\%  & 89.77\% & 93.23\%  \\ \hline %91.55
        \end{tabular}
        \end{adjustbox}
\end{center}
\end{small}
\vspace{0.05 in}
\caption{Comparison of accuracy for different model architectures and datasets}
\label{tab:results}
\vspace{-0.25 in}
\end{table}

\subsection{Using PADs to identify out-of-distribution data} \label{subsec:outofdist}

\begin{figure*}[t]
\begin{center}
    \begin{minipage}[t]{0.3\textwidth}
    \centering
      \includegraphics[width=0.99\textwidth, height = 1.25 in]{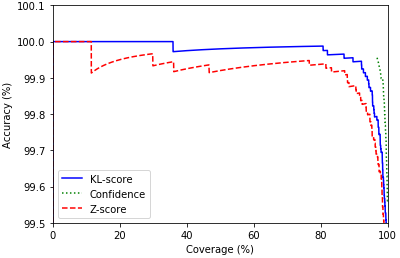}
     \small{\caption{MNIST (MA3): \\Coverage vs. Accuracy\label{fig:mnist_fmnist}}} 
      
    \end{minipage}
     \begin{minipage}[t]{0.44\textwidth}
     \centering
          \includegraphics[width=0.80\textwidth]{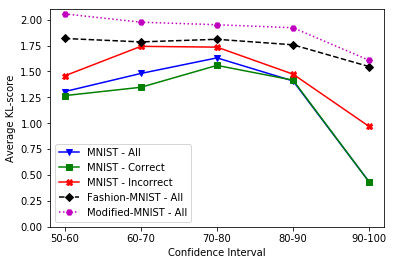}
          \small{\caption{MNIST (MA3):KL-score vs. softmax confidence values, for MNIST \& different OOD data. (notMNIST values outside plotted range)\label{fig:mnist_mnist}}}
     \end{minipage}
    \hfill
    \begin{minipage}[t]{0.24\textwidth}
    \vspace{-1.1 in}
    \centering
    \resizebox{\textwidth}{!}{
    %\begin{small}
    \begin{tabular}{c c c c} 
                    \hline
                    Dataset & S1 & S2 & S3 \\
                    \hline 
                    %\textbf{Ideal-OOD} & 100.00\% & 100.00\% & 100\% \\
                    \textbf{notMNIST}      & 100.00\% & 99.98\% & 99.98\%  \\
                    \textbf{Fashion-MNIST}  & 79.77\% & 97.69\% & 79.27\% \\ 
                    \textbf{Modified-MNIST} & 41.74\% & 98.32\% & 41.71\% \\ 
                    \textbf{MNIST} & 00.68\% & 15.36\% & 00.89\% \\
                    \hline 
        \end{tabular}
   % \end{small}
   }
    \vspace{-0.05 in}
    \begin{small}
    \captionof{table}{MA3: Rejection Rate for OOD vs. in-Distribution data for 3 different strategies. (Ideally, OOD rejection rate= 100\%.)}
    \label{tab:ood_coverage}
    \end{small}
    \end{minipage}
\end{center}
\vspace{-0.3in}
\end{figure*}

% with configuration MA3. If $\alpha$=0.95, $\beta$=0.65, Strategy 1 (S1) := Confidence $\leq\alpha$, Strategy2 (S2) := KL-score $\geq \beta$, Strategy 3 (S3) := Confidence $\leq \alpha$ AND KL-score $\geq\beta$.

Given PAD's intrinsic characterization of the \emph{typical} neural activity for each class, we now demonstrate its use in reliably identifying \emph{out-of-distribution data} (OOD). We used \emph{Fashion-MNIST}, \emph{notMNIST} and  \emph{Modified-MNIST} as exemplars of OOD data, injecting their samples into model MA3 (Table~\ref{tab:datasetsmodels}) that has been trained on the MNIST dataset. For the in-distribution MNIST data, Figure~\ref{fig:mnist_mnist} plots the average KL-score values for samples, categorized by the confidence value produced at the output (softmax) layer. Plots are generated separately for the entire MNIST dataset (MNIST-All), as well as the test samples that are correctly or incorrectly classified (MNIST-Correct \& MNIST-Incorrect, respectively). We see that even when the classifier is highly confident (confidence values $\in \{90,100\}$\%), the KL-score for incorrect samples is more than double (0.97) that of comparable correct samples. Clearly, high KL-scores can help identify incorrect classification attempts.

This trend is further borne out when MA3 is used to classify OOD samples. For both Fashion-MNIST and Modified-MNIST samples, the average KL-score is 3-4 times larger than that obtained for in-distribution samples, even though many OOD samples are associated with high softmax confidence values. To further quantify this, we evaluate 3 different OOD-identification strategies: (i) \emph{S1:} this filters out samples whose softmax confidence is below a threshold $\alpha$ (=0.95 in the MNIST experiments); (ii) \emph{S2}: this PAD-based strategy filters out samples whose KL-score exceeds a threshold $\beta$ (=0.65 in our experiments), and (iii) \emph{S3}: this hybrid strategy filters out only those samples whose $confidence< \alpha \; \mbox{AND} \; KL-score \ge beta$. Table~\ref{tab:ood_coverage} plots the percentage of rejected samples for all 3 strategies. We see that the pure confidence-based \emph{S1} strategy is effective only when the data is completely different (notMNIST), but performs poorly (rejection rate $\sim$40\%) when the OOD dataset has some similarities (Modified-MNIST). In contrast, strategy \emph{S2} can reject the vast majority of OOD samples, but at the cost of a higher rejection rate for in-distribution (MNIST) samples. By combining both predicates, strategy \emph{S3} achieves both higher OOD and low in-distribution rejection rates.

\section{Related Work}\label{sec:relatedwork}
%In this section, we will discuss related work under 2 areas which are of interest to our work. 

\textbf{Hidden-Layers of Deep Neural Networks:} Researchers have explored the interpretability of Convolutional Neural Networks (CNNs) by analyzing their hidden layers~\cite{Yosinski2014,Zeiler2014}. They have suggested that DNNs tend to learn general features such as Gabon filters or Color blobs in the first few layers, while deeper layers learn more dataset-specific features. In an interesting study, Alain \& Bengio \cite{Alain2017} discuss the possibility of creating separately trained linear classifiers aka "probes" using parameters of hidden layers. Similar to our study, they reported that linear separability (and thus classification accuracy) increases as we go deeper in the network. However, their approach requires training a  separate classifiers. Another study \cite{Bau2017} proposes using the alignment of individual hidden units of a CNN to quantify model interpretability. \cite{Raghu2017} studies class specific information in hidden layers of CNNs using Singular Vector Canonical Correlation Analysis (SVCCA). Our methodology has conceptual overlap with ~\cite{Yeh2018, Raghu2017, Alain2017}. However, we believe that PAD provides a novel, generalized construct with multiple uses (unlike~\cite{Alain2017} -- focused purely on classification inference) and defines useful statistical measures on the underlying activation distributions. 

%technique considers hidden-unit activations to obtain a PADs which have many use cases while \cite{Alain2017} trained separate classifiers using a set of parameters to make classifications. These techniques are different even in terms of the preliminary intuition and hypothesis behind them.

% In addition, according to the way they use hidden-layer parameters (not only activations), authors haven't been able to clearly demonstrate a use case for adopting the model except for debugging.

\textbf{Model Pruning:} Researchers have proposed different model pruning strategies (e.g.,~\cite{Lecun1990, Yao2017,Molchanov2017,Anwar2017}) that utilize various properties of hidden layers --e.g., weight-based pruning of convolutional filters or entire nodes. PAD, on the other hand, models a neuron's activation values on a per-class basis, and applies statistical aggregation across multiple neurons, as a means to identify class-specific activation patterns.

\textbf{Uncertainty in DNNs:} Bayesian Neural Networks have been discussed thoroughly in literature as a mathematically grounded way of modelling neural network uncertainty~\cite{Neal1995,Williams1997,MacKay1992}. Recently, there has been a shift towards modelling uncertainty using Bayesian Inference \cite{Herzog2013,Nuzzo2014}. Variational Inference (VI) based Bayesian techniques have been proposed \cite{Lobato2015,Graves2011} even though their validity has been questioned in subsequent research \cite{Ritter2018,Osband2018}. Such Bayesian techniques have higher computational complexity, in both training and inference, and are not yet fully supported in mainstream deep learning libraries. Gal \textit{et. al.} \cite{Gal2016,Gal2016v2} have suggested that Dropout \cite{Srivastava2014} can be utilized to provide Bayesian approximations in DNNs. An alternative technique based on batch normalization was proposed by Teye \textit{et. al.} \cite{Teye2018} which has similar traits to that of \cite{Gal2016v2}. Both these techniques rely on a specific regularization technique (both batch normalization \& dropout in CNNs have associated problems~\cite{Gal2016v3,Springenberg2015,Li2018}). They also require multiple stochastic passes (using the same test sample) to derive an uncertainty measures and are thus not suitable for real-time applications. An ensemble approach for non-Bayesian uncertainty modelling, proposed in  \cite{Lakshminarayanan2017}, requires the use of several DNNs for both training and inferencing.  Another interesting work, \cite{Sensoy2018} employs Dempster-Shafer theory (a generalization of bayesian logic \cite{Gordon1984}) to model uncertainty by adding an additional ``uncertainty class" to the output layer--this method requires changes in training (including loss function and logits). In contrast to several of these approaches, PADs requires no modifications to training, does not employ an explicit Bayesian framework and instead uses low dimensional statistics over the activation values of hidden layer units to distinguish between classes.

\section{Conclusion}\label{sec:conclusion}
We have proposed a novel and intuitive technique, called PAD, to capture class separability in DNNs using the activation values of hidden layer units. Intuitively, PAD leverages on the collective cross-class discrimination capability of all neurons in a hidden layer, provides greater expressivity than available purely at the output layer. As exemplars of PAD's utility, we have demonstrated its use for (a) capturing predictive uncertainty in classification; (b) obtaining highly accurate inferences early, without fully executing a DNN; and (c) filtering out out-of-distribution samples.

We believe that PADs provide a promising representation that can form the basis for interesting future work. For example, (a) PADs may need to be modified to be applicable to other tasks (e.g., regression) beyond just classification, and (b) PADs may provide a mechanism for \emph{class-aware} model compression \& pruning (e.g., by selectively discarding neurons that fire across multiple classes and thus are less discriminative).

\bibliography{7.references.bib}
\bibliographystyle{ACM-Reference-Format}
\end{document}